\documentclass[conference]{IEEEtran}
\IEEEoverridecommandlockouts
\usepackage{cite}
\usepackage{amsmath,amssymb,amsfonts}
\usepackage{graphicx}
\usepackage{textcomp}
\usepackage{xcolor}
\def\BibTeX{{\rm B\kern-.05em{\sc i\kern-.025em b}\kern-.08em
    T\kern-.1667em\lower.7ex\hbox{E}\kern-.125emX}}

\usepackage{algorithm}
\usepackage{algorithmicx}  
\usepackage[noend]{algpseudocode}
\usepackage{amsfonts}
\usepackage{amsmath}
\usepackage{amsthm}
\usepackage{bm}
\usepackage{booktabs}
\usepackage{diagbox}
\usepackage{dsfont} 
\let\labelindent\relax  
\usepackage{enumitem}
\usepackage{etoolbox}
\usepackage{hyperref}  
\usepackage{mathtools}
\usepackage{multirow}
\usepackage{nicefrac}
\usepackage[final]{pdfpages}
\usepackage{siunitx}
\usepackage{stfloats}  
\usepackage[caption=false]{subfig}
\usepackage{tabu}
\usepackage{threeparttable}  
\usepackage{tikz}
\usepackage{xcolor}
\usepackage{xfrac}

\usepackage{pgfplots}   
\pgfplotsset{compat=newest}
\usepgfplotslibrary{groupplots}

\graphicspath{{images/}}
\hypersetup{colorlinks=false}

\newcommand{\method}[1]{\textsc{#1}}

\definecolor{githubColor}{HTML}{2EA44F}
\newcommand{\gitref}[2]{\href{#1}{\color{githubColor}{#2}}}%

\definecolor{newGray}{HTML}{808080}
\definecolor{colorCircle}{HTML}{0072BD}

\definecolor{colorRect}{HTML}{D95319}

\newcolumntype{O}[1]{S[detect-weight, mode=text, table-format=#1]}

\renewcommand{\bfseries}{\fontseries{b}\selectfont} 
\robustify\bfseries             
\newrobustcmd{\B}{\bfseries}

\newcommand\copyrighttext{\footnotesize \textcopyright~2023 IEEE. Personal use of this material is permitted. Permission from IEEE must be obtained for all other uses, in any current or future media, including reprinting/republishing this material for advertising or promotional purposes, creating new collective works, for resale or redistribution to servers or lists, or reuse of any copyrighted component of this work in other works.
DOI: \href{https://doi.org/10.1109/IV55152.2023.10186703}{10.1109/IV55152.2023.10186703}
}

\newcommand\copyrightnotice{%
    \begin{tikzpicture}[remember picture,overlay]%
 	\node[anchor=south, xshift=-0pt, yshift=20pt] at (current page.south)%
 	{\fbox{\parbox{\dimexpr\textwidth-\fboxsep-\fboxrule\relax}{\copyrighttext}}};%
 	\end{tikzpicture}%
}

\newtheoremstyle{tstyle}
  {}
  {}
  {\itshape}
  {}
  {\bfseries}
  {.}
  { }
  {\thmname{#1}\thmnumber{ #2}\thmnote{ (#3)}}%
\theoremstyle{tstyle}

\newcommand{\realnumbers}{\mathbb{R}}

\newcommand{\mbeq}{\overset{!}{=}}

\newcommand{\mat}[1]{\boldsymbol{#1}}
\newcommand{\quat}[1]{\mathrm{#1}}

\renewcommand{\vec}[1]{\boldsymbol{#1}}

\newcommand{\inv}{^\text{\rmfamily \textup{-1}}}
\newcommand{\mneg}{^\text{\rmfamily \textup{-}}}
\newcommand{\mpos}{^\text{\rmfamily \textup{+}}}
\newcommand{\norm}[1]{\left\lVert#1\right\rVert}
\newcommand{\trans}{^\text{\rmfamily \textup{T}}}

\newcommand{\pmat}[1]{\left[#1\right]\mpos}
\newcommand{\nmat}[1]{\left[#1\right]\mneg}

\newcommand{\conj}{^\text{\rmfamily \textup{*}}}

\newcommand{\vectorize}{\operatorname{vec}}

\newcommand{\sol}[1]{\hat{#1}}

\newcommand{\pspace}{\,}  

\title{
Extrinsic Infrastructure Calibration Using\\the Hand-Eye Robot-World Formulation
\thanks{This work was financially supported by the Federal Ministry of Education and Research (BMBF) (project UNICARagil, FKZ\,16EMO0290) and the State Ministry of Economic Affairs Baden-Württemberg (project U-Shift\,II, AZ\,3-433.62-DLR/60).}
}

\author{
\IEEEauthorblockN{
Markus Horn\IEEEauthorrefmark{1},
Thomas Wodtko\IEEEauthorrefmark{1},
Michael Buchholz, and
Klaus Dietmayer
}
\IEEEauthorblockA{
\textit{Institute of Measurement, Control and Microtechnology} \\
\textit{Ulm University}, Germany \\
{\tt \{firstname\}.\{lastname\}@uni-ulm.de}
}
\thanks{\IEEEauthorrefmark{1} \textit{Markus Horn and Thomas Wodtko are co-first authors. Corresponding author: Markus Horn.}}
}

\begin{document}


\maketitle

\begin{abstract}
We propose a certifiably globally optimal approach for solving the hand-eye robot-world problem supporting multiple sensors and targets at once.
Further, we leverage this formulation for estimating a geo-referenced calibration of infrastructure sensors.
Since vehicle motion recorded by infrastructure sensors is mostly planar, obtaining a unique solution for the respective hand-eye robot-world problem is unfeasible without incorporating additional knowledge.
Hence, we extend our proposed method to include a-priori knowledge, i.e., the translation norm of calibration targets, to yield a unique solution.
Our approach achieves state-of-the-art results on simulated and real-world data.
Especially on real-world intersection data, our approach utilizing the translation norm is the only method providing accurate results.
\end{abstract}


\section{Introduction}

\copyrightnotice%
With an increasing amount of intelligent vehicles and infrastructures within urban and rural environments, more and more sensors are required to tackle the accompanying challenges~\cite{buchholz2022handling}.
To enable accurate processing of data acquired by these sensors, sensor calibration is mandatory.
While intrinsic information can often be obtained within the factory, the extrinsic calibration procedure must be appropriate to the given circumstances of the respective work environment~\cite{datondji2016survey}.
For example, calibrating two cameras with overlapping fields of view (FOVs) on a laboratory setup allows for various calibration methods.
In contrast, infrastructure sensors are often mounted high above the ground and can have a limited FOV.
Additionally, strict conditions must be observed, as, for example, roads may not be closed for calibration purposes.
Hence, the task of extrinsically calibrating sensors has to be addressed according to the respective demands.

\begin{figure}
    \centering
    \resizebox{0.9\columnwidth}{!}{%
		\input{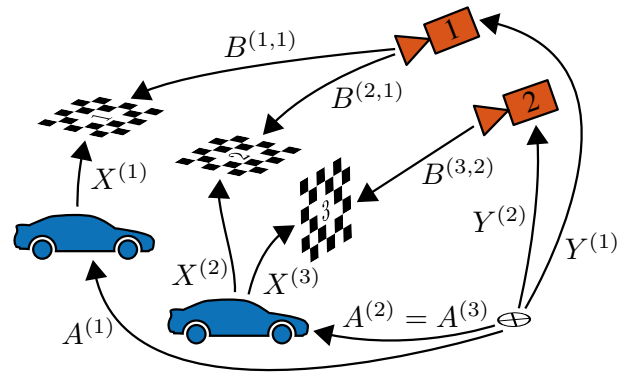}%
	}
    \caption{
    An exemplary scenario for multi-sensor multi-target calibration at a single measurement step $k$ is shown. 
    Two CAVs with known global pose carry multiple, rigidly mounted checkerboard targets detected by two stationary infrastructure cameras.
    For target $t$ and sensor $s$, $A^{(t)}$ represents the geo-referenced pose of the vehicle on which $t$ is mounted and $B^{(t,s)}$ is the detection of $t$ by $s$.
    The desired calibrations are $X^{(t)}$ and $Y^{(s)}$.
    Not all measurements must be simultaneous; only the corresponding $A_k^{(t)}$ must be available for a detection $B_k^{(t,s)}$ at step $k$.
    }
    \label{fig:overview_graph}
\end{figure}

Most approaches for infrastructure calibration are based on static features of the surrounding environment or features of dynamic objects like vehicles.
However, especially for initial calibration, it is also reasonable to use specific calibration targets as long as the calibration does not interfere with traffic and the process is fully automated.
As cameras are among the most commonly used sensors for infrastructures, many calibration procedures are limited to those.
However, since other sensor types, such as lidar, are also used, approaches applicable for multiple sensor types can facilitate the overall calibration process for larger setups.
With a growing number of connected autonomous vehicles (CAVs), a geo-referenced calibration is crucial for fusing data of CAVs and infrastructure sensors.
Thus, a global reference, provided by, e.g., a calibration vehicle, is required.
For this, we propose to mount a sensor-specific calibration target, e.g., a checkerboard for cameras, on a vehicle with a global localization and estimate the sensor and target calibrations.
If the transformation from target to vehicle is known, methods like Perspective-n-Point (PnP) \cite{marchand2016pose} can be used for estimating the sensor calibration.
Unfortunately, accurately measuring the target transformation is time-consuming and requires special equipment, e.g., a coordinate measuring machine like a laser rangefinder.
Hence, we implement this problem using the hand-eye robot-world (HERW) formulation.
Usually, HERW calibration is used to estimate the gripper-to-sensor (hand-eye) and base-to-target (robot-world) calibrations for a moving sensor and a stationary target.
In contrast, the target is moving and the sensor is stationary in our case.
Moreover, infrastructure sensors are often calibrated separately, i.e., each sensor is considered on its own using different calibration features or targets.
In contrast, simultaneously calibrating multiple sensors and targets allows for considering co-dependent errors during optimization.
Hence, we propose a novel multi-sensor multi-target formulation for HERW calibration, as shown in Fig.~\ref{fig:overview_graph}.

This method can obtain geo-referenced sensor calibrations $Y^{(s)}$.
Using specific calibration targets mounted on CAVs facilitates the detection task and allows the approach to be easily adjusted for different sensor types.
Since the target poses $X^{(t)}$ are also estimated, measuring them beforehand is unnecessary.
The vehicle origin for which the global vehicle poses $A^{(t)}$ are given, in our case, the center of the rear axis, does not affect the result of $Y^{(s)}$.

Since roads often have a low change of slope, vehicle motion observed by infrastructure sensors is mainly limited to a flat surface.
Further, if the surrounding area may not be closed for calibration purposes, vehicles must also obey traffic rules, i.e., follow the course of the road.
However, as shown in~\cite{tsai1988real}, at least three measurements with non-colinear inter-pose rotation axes are necessary to solve the HERW problem uniquely.
To enable a calibration, even if it is technically unfeasible to fulfill these requirements, we propose a formulation that allows to integrate additional information into the calibration, i.e., the translation norm of a transformation.
Further, when calibrating multiple sensors simultaneously with our proposed method, these requirements must only hold for the collective amount of data instead of for the individual data of each sensor.

Summarizing our work, we propose
\begin{itemize}[topsep=0pt]
    \item a novel Quadratically Constrained Quadratic Programming (QCQP) formulation for HERW calibration of multiple sensors and targets;
    \item constraints for integrating required a-priori knowledge;
    \item a procedure for geo-referenced infrastructure calibration based on this formulation; and 
    \item a Python-based open-source library available at \gitref{https://github.com/uulm-mrm/excalibur}{https://github.com/uulm-mrm/excalibur}.
\end{itemize}

\section{Related Work}
\label{sec:relatedWork}

\subsection{Hand-Eye Robot-World Calibration}
\label{sec:relWorkHandEye}

A detailed analysis of the HERW problem is presented in~\cite{zhuang1994simultaneous}.
They derive that at least three measurements with non-colinear inter-pose rotation axes are required for a unique solution.
This implies that the HERW calibration problem cannot be solved for planar-only poses without additional information.
We address this issue later in Section~\ref{sec:planarHandling}.

For implementing the HERW problem, different transformation representations are proposed in the literature: quaternions and transformation vectors, homogeneous matrices (HMs), or unit dual quaternions (DQs) are the most common.
However, no superior representation suitable for all scenarios was found in~\cite{tabb2017solving, heller2014handeye}.
Since we have already derived an efficient unit DQ formulation for hand-eye calibration in~\cite{horn2021online}, we extend this formulation for HERW calibration.

A common choice as a cost function for camera calibration is the reprojection error~\cite{tabb2017solving, li2018eye, ali2019methods}.
Since an initial calibration is required for most of these approaches, an inaccurate initial solution can lead to non-convergence~\cite{ali2019methods}.
Furthermore, the reprojection error is unsuitable for other sensors without available correspondences.
Therefore, we focus on directly optimizing the transformation cycle error.

For quaternions and HMs, the rotations of $X$ and $Y$ can be optimized separately from the translations, which are optimized in a second step~\cite{zhuang1994simultaneous, dornaika1998simultaneous, shah2013solving, tabb2017solving, wang2022accurate}.
Even though this can yield lower rotation errors, it can also lead to higher translation and reprojection errors due to error propagation from the first step~\cite{dornaika1998simultaneous, tabb2017solving}.
This is problematic for subsequent detection and reconstruction tasks, suggesting a simultaneous estimation of rotations and translations.

For a simultaneous estimation, the Kronecker product formulation for HMs~\cite{li2010simultaneous, li2018simultaneous} or the matrix formulation of unit DQs~\cite{li2010simultaneous} is commonly used to enable a linear formulation with a closed-form solution.
However, the performance of linear methods degrades in the presence of noise~\cite{dornaika1998simultaneous}.
Furthermore, these methods do not enforce valid HMs or unit DQs when solving the problem but instead obtain a valid transformation from their solution afterward, which can lead to errors~\cite{shah2013solving}.

An early approach with a non-linear cost function based on HMs, enforcing valid solutions using weighted penalty terms, was presented in~\cite{dornaika1998simultaneous}.
However, they still require an initial solution and an appropriate weight selection.
Instead, constraints are better suited for enforcing valid solutions during optimization, as proposed in~\cite{zhao2019simultaneous, heller2014handeye}.
Both approaches yield globally optimal solutions but suffer from high run times of multiple seconds or minutes, even for only a few measurements and a single sensor-target pair.
In contrast, our proposed method allows for a larger number of measurements with multiple sensors and targets while still efficiently obtaining certifiably globally optimal solutions.
Further, multiple sensors are only supported by~\cite{tabb2017solving, wang2022accurate}.
To the best of our knowledge, our proposed method is the first to support multiple sensors and targets.

\subsection{Infrastructure Calibration}

A large number of approaches for infrastructure calibration are based on vanishing points~\cite{kanhere2010taxonomy} that are extracted from lane markings~\cite{fung2003camera}, vehicle motion and edges~\cite{dubska2014automatic, dubska2015fully}, or vehicle box detections~\cite{sochor2017traffic}.
Another class of approaches uses landmark detections on vehicles~\cite{bhardwaj2018autocalib, bartl2021automatic}.
All mentioned approaches focus on calibration in the usual work environment of infrastructure cameras without specific calibration targets.
However, in contrast to our method, they are unsuitable for other sensor types, multiple sensors without overlapping FOVs, or geo-referenced calibration.
Since calibration is not performed frequently, relying on calibration targets is still viable if the procedure is fully automated.

Other approaches for geo-referenced calibration of sensors with non-overlapping FOVs were presented in~\cite{mueller2019laci, tsaregorodtsev2022extrinsic}.
However, these methods are designed for only a single sensor of a specific type, i.e., lidar or camera, whereas our approach can be applied to multiple sensors of different types as long as a suitable calibration target is available.

\section{Foundations}

In this work, unit DQs are used to formulate the HERW problem.
For this, DQs and the corresponding QCQP structure are introduced in this section.

Transformations are denoted as uppercase letters $T$ and are generally referred to as functions, meaning no representation is implicitly specified.
A transformation is always directed frame-forward.
The composition of two transformations is denoted by $T_a \circ T_b$.
Vectors and matrices are denoted by lowercase and uppercase bold letters $\vec{x}$ and $\mat{X}$, respectively.
(Dual) quaternions are denoted as upright letters~$\quat{q}$.

\subsection{Dual Quaternions}

In general, a DQ $\quat{q} = \quat{r} + \epsilon \, \quat{d}$ consists of the real part $\quat{r}$ and the dual part $\quat{d}$ with the dual unit $\epsilon$ with $\epsilon^2 = 0$~\cite{mccarthy1990introduction}.
For representing rigid transformations, the real part is a unit rotation quaternion, and the dual part can be calculated using the translation vector $\vec{t} \in \realnumbers^3$ as
\begin{align}
\label{eq:dualPart}
    \quat{d} = \tfrac{1}{2} \quat{t} \quat{r} \pspace , \quad \text{with\ } \quat{t} = \left(0,\vec{t}\right) \pspace .
\end{align}
This implies that a transformation DQ must be unit, i.e., satisfying the constraints
\begin{align}
\label{eq:unitDualQuat}
    \norm{\quat{r}}^2 = \quat{r} \, \quat{r}\conj = 1
    \quad \text{and} \quad
    \quat{r} \, \quat{d}\conj + \quat{d}  \, \quat{r}\conj = 0 \pspace ,
\end{align}
as derived in~\cite{horn2021online}.
Further, the unit DQs $\quat{q}$ and $-\quat{q}$ represent the same transformation.
This must be considered when deriving the optimization problem in Section~\ref{sec:method}.
A DQ can also be represented in vectorized form $\vec{q} = \vectorize{(\quat{q})}$ with $\vec{q} \in \realnumbers^8$.
Furthermore, a DQ multiplication $\quat{q}_a \, \quat{q}_b$ can be represented as a matrix-vector product
\begin{align}
\label{eq:dqMultMatrix}
    \vectorize{(\quat{q}_a \, \quat{q}_b)} =
    \pmat{\mat{Q}_a} \vec{q}_b =
    \nmat{\mat{Q}_b} \vec{q}_a
\end{align}
where $\pmat{\mat{Q}_a}$ and $\nmat{\mat{Q}_b}$ denote the left and right DQ matrix representations, respectively~\cite{mccarthy1990introduction}.
Using the matrix representation, the constraints~\eqref{eq:unitDualQuat} can be expressed as
\begin{subequations}
\label{eq:constraintsMatrices}
\begin{align}
    1 + \vec{q}\trans \, \mat{P}_r \, \vec{q} = 0
    \quad &\text{and} \quad
    \vec{q}\trans \, \mat{P}_d \, \vec{q} = 0 \pspace , \\
    \text{with }
    \mat{P}_r :=
    \begin{bmatrix}
    -\mat{I}_{4} & \mat{0}_{4} \\
    \mat{0}_{4} & \mat{0}_{4} \\
    \end{bmatrix}
    \quad &\text{and} \quad
    \mat{P}_d :=
    \begin{bmatrix}
    \mat{0}_{4} & \mat{I}_{4} \\
    \mat{I}_{4} & \mat{0}_{4} \\
    \end{bmatrix} \pspace .
\end{align}
\end{subequations}
Next, this is used to formulate the optimization problem.

\subsection{QCQP for Dual Quaternions}
\label{sec:qcqp}

Given a positive semi-definite matrix $\mat{Q}$, a purely quadratic cost or objective function $J(\vec{z}) = \vec{z}\trans \mat{Q} \vec{z}$ can be constructed.
If $\vec{z}$ represents a single vectorized unit DQ, an optimization problem with the objective function $J$ is defined by
\begin{subequations}
\label{eq:primalProb}
\begin{alignat}{2}
    & \!\min_{\vec{z}} & ~ & J(\vec{z}) := \vec{z}\trans \, \mat{Q} \, \vec{z}\\
    & \text{w.r.t.}	 & ~ & \vec{g}(\vec{z}) :=
        \begin{pmatrix} 
        1 + \vec{z}\trans \, \mat{P}_r \, \vec{z} \\
        \vec{z}\trans \, \mat{P}_d \, \vec{z}
        \end{pmatrix} \mbeq \vec{0} \pspace .
\end{alignat}
\end{subequations}
The constraints \eqref{eq:constraintsMatrices} are required to ensure a valid unit DQ during optimization.
The respective Lagrangian dual problem can be derived analogously to~\cite{horn2021online,wodtko2021globally}.
Given the Lagrangian function
\begin{align}
    L(\vec{z},\vec{\lambda}) = \vec{z}\trans\mat{Z}(\vec{\lambda})\vec{z} + \lambda_1  \pspace ,
\end{align}
with $\mat{Z}(\vec{\lambda}) := \mat{Q} + \lambda_1 \mat{P}_{r} + \lambda_2 \mat{P}_{d}$, the dual problem is
\begin{subequations}
\label{eq:dualProb}
\begin{alignat}{2}
&\!\max_{\vec{\lambda}}  &\quad& \lambda_1\\
&\text{w.r.t.} &      & \mat{Z}(\vec{\lambda}) \succeq \mat{0}  \pspace .
\end{alignat}
\end{subequations}
It is known from duality theory \cite{boyd2004convex} that the Lagrangian dual problem is always convex, and its value at the solution $\sol{\vec{\lambda}}$ provides a lower bound for the primal optimal value $J(\sol{\vec{z}})$.
The so-called duality gap describes the difference between primal and dual optimal values.
If the duality gap is zero, we know that $\sol{\vec{z}}$ is a globally optimal primal solution.
In the next section, we describe the procedure for recovering the primal solution $\sol{\vec{z}}$ from the dual solution $\sol{\vec{\lambda}}$.

For later use, the primal and dual problems \eqref{eq:primalProb} and \eqref{eq:dualProb} can be extended so that multiple DQs can be optimized.
Given an extended optimization variable $\vec{z} = \begin{bmatrix} \vec{q}_1\trans & \dots & \vec{q}_n\trans \end{bmatrix}\trans$ containing $n$ vectorized DQs, the constraints \eqref{eq:constraintsMatrices} must be enforced separately for each DQ.
For this, the constraint matrices are replicated as described later in Section~\ref{sec:multiSensorMultiTarget}.
Finally, the cost matrix $\mat{Q}$ must be constructed according to the respective problem to fit the dimension of $\vec{z}$.

\subsection{Recover Primal from Dual Solution}
\label{sec:recovery}

As shown in~\cite{horn2021online}, a certifiably globally optimal primal solution $\sol{\vec{z}}$ can be recovered from the null space of $\mat{Z}(\sol{\vec{\lambda}})$.
If the null space has dimension 1, the solution is recovered by normalizing the null space vector using the real part constraint of \eqref{eq:unitDualQuat}.
If the null space has dimension 2, the solution is recovered using the equations derived in \cite{daniilidis1999hand}.
For a larger null space, the non-convex primal problem is optimized directly, with the normalized null space vector corresponding to the largest eigenvalue as initial solution.
After recovering the primal solution, the duality gap must be checked.
Only if the duality gap is zero, it is verified that the recovered solution is globally optimal.

\section{Problem Definition}
\label{sec:problem}

The objective of HERW calibration with multiple targets $t = 1, \dots, m_X$ and multiple sensors $s = 1, \dots, m_Y$ is to estimate the respective calibrations $X^{(t)}$ and $Y^{(s)}$, as shown in Fig.~\ref{fig:overview_graph}.
Hence, in total $m = m_X + m_Y$ transformations must be estimated simultaneously.
At each measurement step $k = 1, \dots, n$, there can be detections $B_k^{(t, s)}$ of target $t$ by sensor $s$.
The respective pose of the reference frame for target $t$, in our case the global vehicle pose, is given by $A_k^{(t)}$.
This is summarized in the general transformation equation
\begin{align}
\label{eq:herwMulti}
    A_{k}^{(t)} \circ X^{(t)} = Y^{(s)} \circ B_{k}^{(t,s)} \pspace .
\end{align}
The goal is to minimize the error of \eqref{eq:herwMulti} over all available measurements and transformations.
It shall be mentioned that not for every $k$, poses and detections must be available for every target-sensor pair.

For our main application, infrastructure calibration, only the sensor calibration $Y^{(s)}$ is of interest.
Nevertheless, the relative target poses $X^{(t)}$ are required for the concept described in Section~\ref{sec:procedure}.
Since measuring $X^{(t)}$ by hand is time-consuming, simultaneously estimating all transformations is preferred.

\section{Method}
\label{sec:method}

In this section, a unit DQ-based formulation for HERW calibration is presented.
First, a single sensor-target pair is used to derive the underlying optimization structure with all necessary constraints.
Consecutively, the problem formulation is extended to allow for calibrating multiple sensors and targets simultaneously.
Lastly, constraints are derived to allow for integrating a-priori knowledge into the calibration procedure.

\subsection{Hand-Eye Robot-World Optimization Problem}

First, calibration with a single sensor and a single target is presented.
In this case, HERW calibration is described by
\begin{align}
\label{eq:herw}
    A_k \circ X = Y \circ B_k \pspace ,
\end{align}
where $X$ and $Y$ represent the desired calibrations, and $A_k$ and $B_k$ are measurements at step $k$.
Implementing \eqref{eq:herw} for a single step $k$ using DQs and their matrix representation~\eqref{eq:dqMultMatrix} leads to
\begin{subequations}
\begin{alignat}{2}
    & & \quat{q}_{a} \quat{q}_{x} &= \quat{q}_{y} \quat{q}_{b} \\
    \quad \iff \quad & &
    \quat{q}_{x} &= \quat{q}_{a}\inv \quat{q}_{y} \quat{q}_{b} \\
    \quad \iff \quad & &
    \vec{x} &= \pmat{\mat{A}\inv} \nmat{\mat{B}} \vec{y} =: \mat{C} \vec{y} \pspace ,
\end{alignat}
\end{subequations}
with $\mat{C} = \pmat{\mat{A}\inv} \nmat{\mat{B}}$, which can be expressed by
\begin{align}
\label{eq:Mz}
    \mat{M} \vec{z} :=
    \begin{bmatrix}
        \mat{I}_{8} & - \mat{C} \\
    \end{bmatrix}
    \begin{bmatrix}
        \mat{x} \\ \mat{y}
    \end{bmatrix}
    =
    \vec{0} \pspace .
\end{align}
Here, it is crucial to handle the sign ambiguity of unit DQs.
This is further described in Section~\eqref{sec:signAmbiguity}.
The matrix $\mat{M}$ contains all measurements of a single step.
Given $n$ steps, the quadratic cost function $J$ is constructed with $\mat{Q}$ defined by
\begin{align}
    \mat{Q} := \sum_{k=1}^{n} \mat{M}_{k}\trans \mat{M}_{k} \pspace .
\end{align}
As mentioned in Section~\ref{sec:qcqp}, further constraints must be added for ensuring $\vec{x}$ and $\vec{y}$ to represent valid unit DQs.
The corresponding constrant matrices are given by
\begin{subequations}
\label{eq:constraintsMatKron}
\begin{alignat}{3}
    \mat{P}_{r, \quat{x}} &= \mat{E}_{1} \otimes \mat{P}_r \pspace , &\quad
    \mat{P}_{d, \quat{x}} &= \mat{E}_{1} \otimes \mat{P}_d \pspace ,\\
    \mat{P}_{r, \quat{y}} &= \mat{E}_{2} \otimes \mat{P}_r \pspace , &\quad
    \mat{P}_{d, \quat{y}} &= \mat{E}_{2} \otimes \mat{P}_d \pspace ,
\end{alignat}
\end{subequations}
with $\mat{E}_{i} := \vec{e}_i \vec{e}_i\trans$, using the canonical basis vector $\vec{e}_i \in \mathds{R}^2$.
The Kronecker product~$\otimes$ with $\mat{E}_{i}$ places the matrices $\mat{P}_r$ and $\mat{P}_d$ from \eqref{eq:constraintsMatrices} at the locations for the corresponding DQs $\quat{x}$ and $\quat{y}$.
Finally, the Lagrangian dual problem for the HERW formulation is given analogously to \eqref{eq:dualProb} as
\begin{subequations}
\label{eq:dualProbHERW}
\begin{alignat}{2}
    &\!\max_{\vec{\lambda}}  &\quad& \lambda_1 + \lambda_3\\
    &\text{w.r.t.} &      & \mat{Z}(\vec{\lambda}) \succeq \mat{0} \pspace ,
\end{alignat}\\[-35pt]
\begin{align}
    \text{with} \quad
    \mat{Z}(\vec{\lambda}) := \mat{Q} + 
    &\lambda_1 \mat{P}_{r,\quat{x}} + 
    \lambda_2 \mat{P}_{d,\quat{x}} \, + \\
    &\lambda_3 \mat{P}_{r,\quat{y}} + 
    \lambda_4 \mat{P}_{d,\quat{y}} \pspace .
\end{align}
\end{subequations}
The solution to this problem is used to recover a certifiably globally optimal primal solution, as described in Section~\ref{sec:recovery}.

\subsection{Dual Quaternion Sign Ambiguity Handling}
\label{sec:signAmbiguity}

Due to the sign ambiguity of unit DQs, $\quat{q}_{x}$ and $-\quat{q}_{x}$ represent the same transformation.
However, this equality does not hold in the matrix-vector form \eqref{eq:Mz}, since $\vec{x} \neq -\vec{x}$.
This means that for each cycle, depending on the signs of $\quat{q}_{a}$ and $\quat{q}_{b}$, either $\vec{z} = \begin{bmatrix} \vec{x}\trans & \vec{y}\trans \end{bmatrix}\trans$ or $\vec{z}' = \begin{bmatrix} \vec{x}\trans & -\vec{y}\trans \end{bmatrix}\trans$ are the optimal solution.
Therefore, these signs must be selected accordingly for each cycle.

Since switching both signs does not affect the solution, only the sign of $\quat{q}_{b}$ must be selected appropriately in each cycle.
In \cite{heller2014handeye}, this is done by comparing all $2^{n-1}$ combinations and selecting the best result.
However, since this is impractical for a large number of samples,
we propose a novel approach inspired by RANSAC \cite{fischler1981random}.
First, three samples are randomly selected, which are usually sufficient for obtaining a unique solution.
For these samples, only four different sign combinations must be checked.
The combination leading to the lowest costs is then used to select the signs for all other samples.
This process can be repeated with different random samples in case of outliers or noise.
For multiple targets or multiple sensors, as described in the next section, this concept is applied individually for each target-sensor pair.

\subsection{Multiple Sensors and Targets}
\label{sec:multiSensorMultiTarget}

This section extends the previously derived single-sensor single-target calibration to multiple sensors and targets, as described by \eqref{eq:herwMulti}.
By adjusting the derivation for \eqref{eq:Mz}, the extended vector-matrix notation is given by
\begin{subequations}
\begin{align}
    \label{eq:Mkts}
    \mat{M}_{k}^{(t,s)} &:=
    \begin{bmatrix}
        \vec{e}_{t, m_X}\trans \otimes \mat{I}_{8} &
        -\vec{e}_{s, m_Y}\trans \otimes \mat{C}_{k}^{(t,s)} 
    \end{bmatrix} \pspace , \\
    \vec{z} &:= 
    \begin{bmatrix}
        \vec{x}_1\trans & \dots & \vec{x}_{m_X}\trans & 
        \vec{y}_1\trans & \dots & \vec{y}_{m_Y}\trans 
    \end{bmatrix}\trans \pspace ,
\end{align}
\end{subequations}
with the canonical basis vectors $\vec{e}_{i, m_{\ast}} \in \mathds{R}^{m_{\ast}}$, $\mat{M}_{k}^{(t,s)} \in \mathds{R}^{8 \times 8m}$, and $\vec{z} \in \mathds{R}^{8m}$.
The cost matrix $\mat{Q}$ of \eqref{eq:primalProb} is then calculated by
\begin{align}
    \label{eq:QfromMkts}
    \mat{Q} := \sum_{(k, t, s) \in \mathds{D}} {\mat{M}_{k}^{(t,s)}}\trans \mat{M}_{k}^{(t,s)} \pspace ,
\end{align}
with the detection set~$\mathds{D}$ containing tuples of all targets~$t$ detected by sensor~$s$ at step~$k$.
Analogous to \eqref{eq:constraintsMatKron}, the additional constraint matrices are defined by
\begin{subequations}
\label{eq:multiConstraints}
\begin{align}
    \mat{P}_{r,j} = \mat{E}_j \otimes \mat{P}_r
    \quad \text{and} \quad
    \mat{P}_{d,j} = \mat{E}_j \otimes \mat{P}_d \pspace ,
\end{align}
\end{subequations}
with $j = 1, \dots, m$ and $\mat{E}_j \in \mathds{R}^{m \times m}$.
This finally leads to the Lagrangian dual problem
\begin{subequations}
\label{eq:dualProbHERWMulti}
\begin{alignat}{2}
    &\!\max_{\vec{\lambda}}  &\quad& \sum_{j = 1}^{m}\lambda_{2j-1}\\
    &\text{w.r.t.} &      & \mat{Z}(\vec{\lambda}) \succeq \mat{0}  \pspace ,
\end{alignat}\\[-30pt]
\begin{align}
\label{eq:Zmulti}
    \text{with} \quad
    \mat{Z}(\vec{\lambda}) := \mat{Q} + \sum_{j = 1}^{m}\lambda_{2j-1} \mat{P}_{r,j} + \lambda_{2j} \mat{P}_{d,j} \pspace ,
\end{align}
\end{subequations}
for solving the HERW problem simultaneously for multiple sensors and targets.

\subsection{Translation Norm Constraints}
\label{sec:transNormConst}

The measurement requirements mentioned in Section~\ref{sec:relWorkHandEye} must be satisfied for $A_k^{(t)}$ and $B_k^{(t,s)}$ to guarantee a unique solution~\cite{tsai1988real}.
As shown later in Section~\ref{sec:planarHandling}, integrating additional information can mitigate some of these conditions.
Hence, we introduce novel constraints allowing for integrating a-priori translation norm knowledge into the calibration.
From \eqref{eq:dualPart} and \eqref{eq:unitDualQuat}, the norm of the dual part relates to the respective translation norm by
\begin{align}
    \norm{\quat{d}}^2
    = \norm{\tfrac{1}{2} \quat{t} \quat{r}}^2
    = \tfrac{1}{4} \norm{\quat{t}}^2 \pspace .
\end{align}
Given a known translation norm $\norm{\quat{t}} = \alpha$, it follows that
\begin{align}
    \vec{d}\trans \vec{d} = \frac{1}{4} \alpha^2  \pspace .
\end{align}
Thus, the additional constraint
\begin{subequations}
\label{eq:normConstraint}
\begin{gather}
    \frac{1}{4} \alpha_j^2 + \vec{z}\trans \, \mat{P}_{\mathrm{norm}, j} \, \vec{z}= 0 \pspace , \\
    \text{with\ } \mat{P}_{\mathrm{norm}, j} =
    \mat{E}_j \otimes
    \begin{bmatrix}
        \mat{0}_{4} & \mat{0}_{4} \\
        \mat{0}_{4} & -\mat{I}_{4} \\
    \end{bmatrix} \pspace ,
\end{gather}
\end{subequations}
can be used to enforce the translation norm of transformation $j$ during optimization.
This is later applied in Section~\ref{sec:planarHandling} to enable HERW calibration from planar-only pose data.

\section{Calibration Procedure}
\label{sec:procedure}

In this section, our proposed calibration procedure for infrastructure sensors is described.
For better readability, the algorithm is only outlined for a single target.
Hence, the vehicle poses $A_k$ and target detections $B_k^{(s)}$ are used to obtain the target calibration $X$ and sensor calibrations $Y^{(s)}$.

Section~\ref{sec:calibTarget} describes a calibration target for fully automatic camera detection.
Since the vehicle motion $A_k$ is mostly planar, a unique HERW solution can only be acquired by utilizing additional information.
This procedure is described in Section~\ref{sec:planarHandling}.

\subsection{Calibration Target}
\label{sec:calibTarget}

\begin{figure}
    \centering
    \subfloat[Camera 1]{%
        \label{fig:vehicleImage:cam1}%
    	\includegraphics[
    	    width=0.495\columnwidth,
    	    trim={350px 0 100px 250px},
    	    clip
    	]{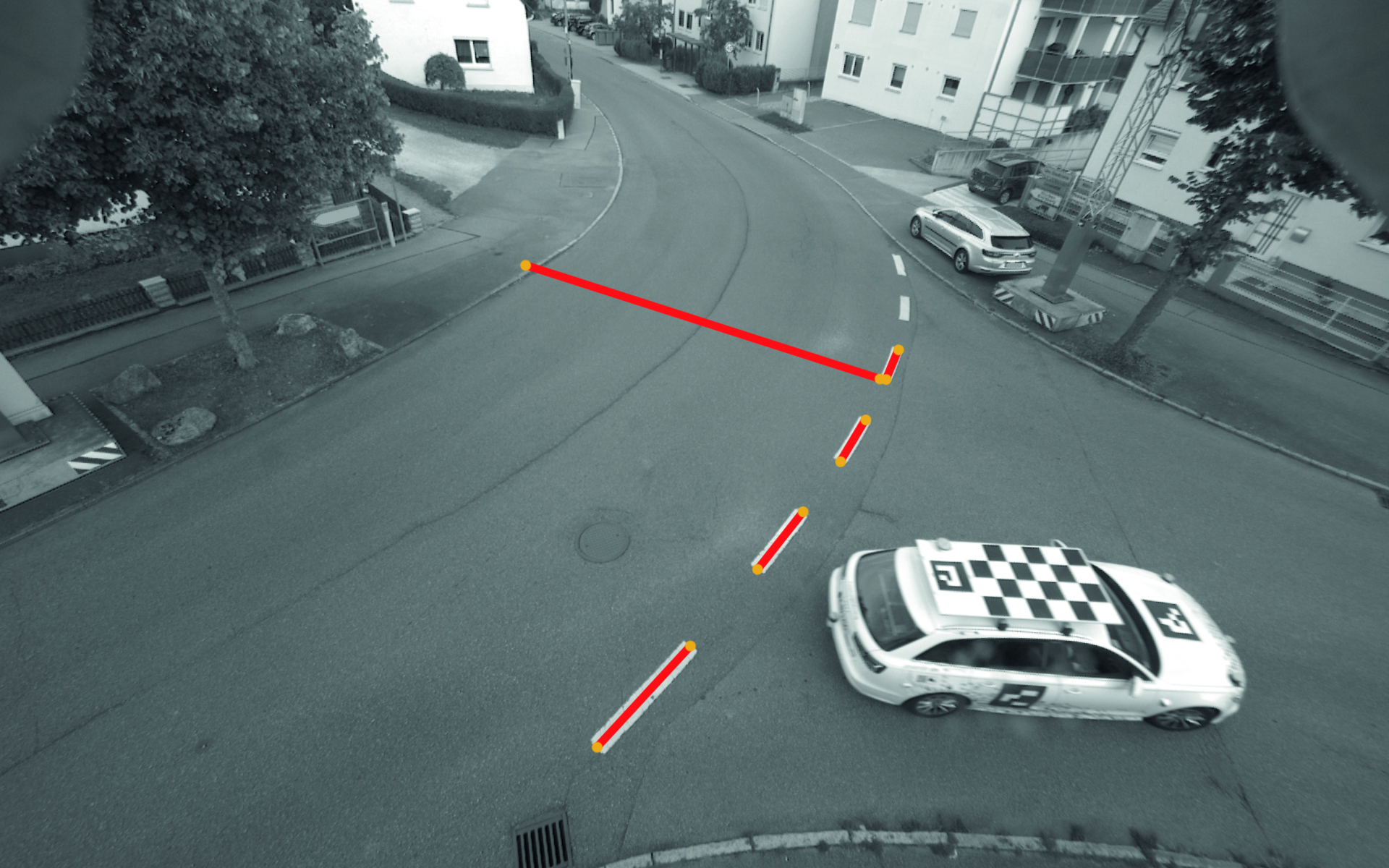}%
    }
    \hfill
    \subfloat[Camera 2]{%
        \label{fig:vehicleImage:cam2}%
    	\includegraphics[
    	    width=0.495\columnwidth,
    	    trim={250px 0 600px 510px},
    	    clip
    	]{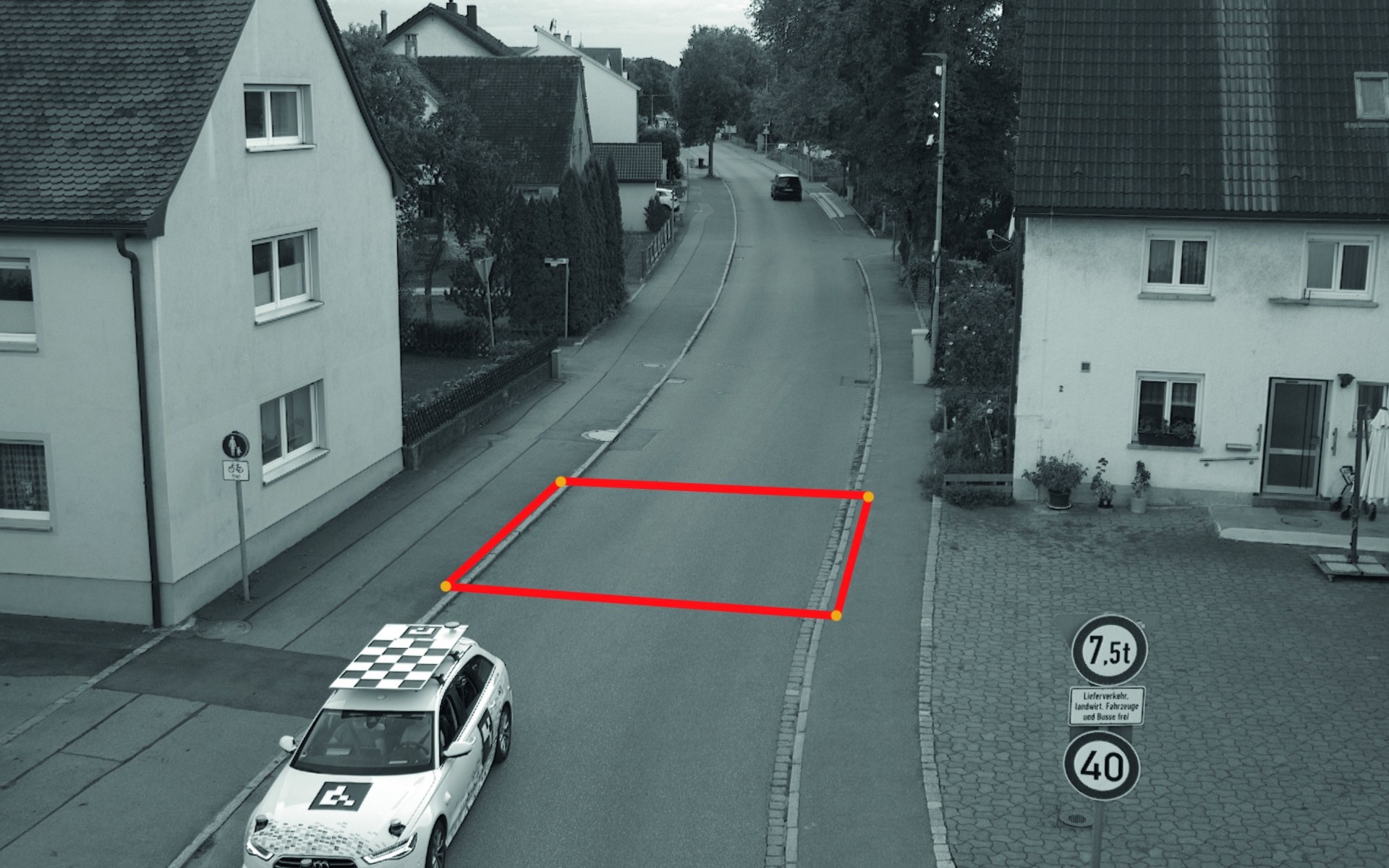}%
    }
    \caption{%
    Cropped camera images of the two infrastructure cameras used for evaluation are shown.
    The calibration vehicle with ArUco markers and checkerboard is visible in both images.
    Further, the reference lines used for quantitatively evaluating the camera calibrations are marked red.
    }
    \label{fig:vehicleImage}
\end{figure}

A combination of multiple ArUco markers~\cite{garrido2014automatic} and a checkerboard~\cite{furgale2013unified} is used as calibration target, shown in Fig.~\ref{fig:vehicleImage}.
The ArUco markers enable quick detection and unique identification of the respective calibration vehicle from all directions in camera images.
Hence, they help to identify a region of interest for the subsequent checkerboard detection.
For the cameras in Section~\ref{sec:realWorldIntersection}, the markers are detected at distances up to \SI{20}{\meter}.
This estimate is then used to detect the checkerboard corners efficiently with sub-pixel accuracy.
Afterward, the checkerboard's pose in camera coordinates is calculated using the intrinsic camera calibration and the known checkerboard dimensions.
The unique orientations of the ArUco markers are finally used to ensure a unique orientation of the checkerboard pose.

\subsection{Translation Correction for Planar-Only Poses}
\label{sec:planarHandling}

\begin{figure}
    \centering
    \resizebox{0.65 \columnwidth}{!}{%
\begingroup%
  \makeatletter%
  \providecommand\color[2][]{%
    \errmessage{(Inkscape) Color is used for the text in Inkscape, but the package 'color.sty' is not loaded}%
    \renewcommand\color[2][]{}%
  }%
  \providecommand\transparent[1]{%
    \errmessage{(Inkscape) Transparency is used (non-zero) for the text in Inkscape, but the package 'transparent.sty' is not loaded}%
    \renewcommand\transparent[1]{}%
  }%
  \providecommand\rotatebox[2]{#2}%
  \newcommand*\fsize{\dimexpr\f@size pt\relax}%
  \newcommand*\lineheight[1]{\fontsize{\fsize}{#1\fsize}\selectfont}%
  \ifx\svgwidth\undefined%
    \setlength{\unitlength}{146.4375bp}%
    \ifx\svgscale\undefined%
      \relax%
    \else%
      \setlength{\unitlength}{\unitlength * \real{\svgscale}}%
    \fi%
  \else%
    \setlength{\unitlength}{\svgwidth}%
  \fi%
  \global\let\svgwidth\undefined%
  \global\let\svgscale\undefined%
  \makeatother%
  \begin{picture}(1,0.68245839)%
    \lineheight{1}%
    \setlength\tabcolsep{0pt}%
    \put(0,0){\includegraphics[width=\unitlength,page=1]{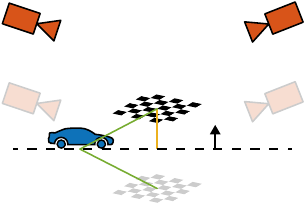}}%
    \put(0.34613572,0.26718054){\makebox(0,0)[lt]{\lineheight{1.25}\smash{\begin{tabular}[t]{l}$\alpha$\end{tabular}}}}%
    \put(0.34613572,0.09304481){\makebox(0,0)[lt]{\lineheight{1.25}\smash{\begin{tabular}[t]{l}$\alpha$\end{tabular}}}}%
    \put(0.52795391,0.22279129){\makebox(0,0)[lt]{\lineheight{1.25}\smash{\begin{tabular}[t]{l}$\gamma$\end{tabular}}}}%
    \put(0.73026376,0.23303457){\makebox(0,0)[lt]{\lineheight{1.25}\smash{\begin{tabular}[t]{l}$\vec{u}_{\text{w}}$\end{tabular}}}}%
  \end{picture}%
\endgroup%
	}
    \caption{
    A vehicle carrying a single checkerboard target and two infrastructure sensors are illustrated.
    The dashed line represents the plane of vehicle motion along with its up vector $\vec{u}_\text{w}$ in world coordinates.
    In this scenario, the HERW formulation with known translation norm $\alpha$ can obtain two possible solutions, which are shifted by $2\gamma$ along the up vector.
    The desired solution (colored) shows the checkerboard above the vehicle, while the other possible solution (transparent) would require it to be below the vehicle.
    }
    \label{fig:twoSolutions}
\end{figure}
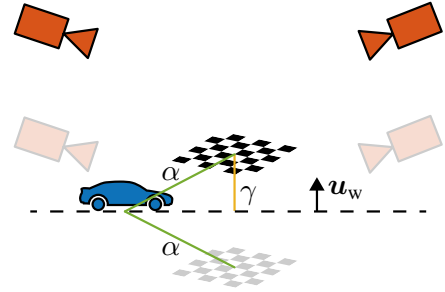

As mentioned in Section~\ref{sec:transNormConst}, no unique HERW solution can be obtained from planar-only poses without additional information.
More precisely, the translation is indefinite along the direction of the motion plane up vector~\cite{tsai1988real}.
By providing only the translation norm of the target calibration $X$ using the constraint described in~\eqref{eq:normConstraint}, the number of valid solutions is reduced to two, as displayed in Fig.~\ref{fig:twoSolutions}.
This norm can be easily obtained using, e.g., a measuring tape.
Since the optimization favors one of both solutions, primarily due to the influence of noise, the solutions $\sol{X}$ and $\sol{Y}^{(s)}$ can be checked and adjusted automatically.
For this, it is assumed w.l.o.g. that the target is mounted above the vehicle origin, i.e., in the positive $z$-direction of the vehicle frame located at the rear axis center.

In both the vehicle and the world frame, the plane up vectors $\vec{u}_{\text{v}}$ and $\vec{u}_{\text{w}}$ are obtained from the positions of $A_k\inv$ and $A_k$, respectively, using Principal Component Analysis.
By projecting the translation vector $\sol{\vec{t}}_X$ of the estimated target calibration $X$ onto $\vec{u}_{\text{v}}$, the orthogonal height offset $\gamma = \vec{u}_{\text{v}}\trans \, \sol{\vec{t}}_X$ is calculated.
If $\gamma$ is positive, the estimated checkerboard pose~$\sol{X}$ is above the vehicle origin and, thus, the desired solution.
Otherwise, $\sol{X}$ and $\sol{Y}^{(s)}$ must be adjusted by
\begin{subequations}
\label{eq:adjustCalib}
\begin{alignat}{3}
    {\sol{X}}'       & = T_{\vec{u}_X} \circ \sol{X}       \pspace , & \quad \text{with } \vec{u}_X & = -2 \, \gamma \, \vec{u}_{\text{v}} \pspace , \\
    \sol{Y}^{(s)}{}' & = T_{\vec{u}_Y} \circ \sol{Y}^{(s)} \pspace , & \quad \text{with } \vec{u}_Y & = -2 \, \gamma \, \vec{u}_{\text{w}} \pspace ,
\end{alignat}
\end{subequations}
where $T_{\vec{u}}$ represents a transformation with translation vector $\vec{u}$ and no rotation.
The entire process for optimizing and obtaining the desired solution is summarized in Algorithm~\ref{alg:infrastructurCalib}.

\begin{algorithm}
\caption{Infrastructure Calibration}
\label{alg:infrastructurCalib}
\hspace*{\algorithmicindent} \textbf{Input:} vehicle poses $A_{k}$, target detections $B_{k}^{(s)}$, \\
\hspace*{\algorithmicindent}\phantom{\textbf{Input:} } target calibration translation norm $\alpha$ \\
\hspace*{\algorithmicindent} \textbf{Output:} estimated calibrations $\sol{X}, \sol{Y}^{(s)}$
\begin{algorithmic}[1]
\Procedure{Calibrate}{$A_{k}$, $B_{k}^{(s)}$, $\alpha$}
    \State Form $\mat{M}_{k}^{(s)}$ from $A_{k}$ and $B_{k}^{(s)}$ \Comment{\eqref{eq:Mkts}}
    \State Form $\mat{Q}$ from all $\mat{M}_{k}^{(s)}$ \Comment{\eqref{eq:QfromMkts}}
    \State Form $\mat{P}_{r,j}$, $\mat{P}_{d,j}$, and $\mat{P}_{\mathrm{norm}, j}$ for $\alpha_j$ \Comment{\eqref{eq:multiConstraints}, \eqref{eq:normConstraint}}
    \State $\sol{X}, \sol{Y}^{(s)} \gets$ Global HERW solution  \Comment{\eqref{eq:dualProbHERWMulti}, Sec.~\ref{sec:recovery}}
    \State $\gamma \gets$ Projection of $\sol{\vec{t}}_X$ onto $\vec{u}_{\text{v}}$  
    \If {$\gamma < 0$}
        \State $\sol{X}, \sol{Y}^{(s)} \gets$ Adjust $\sol{X}$ and $\sol{Y}^{(s)}$ \Comment{\eqref{eq:adjustCalib}}
    \EndIf
    \State \Return $\sol{X}, \sol{Y}^{(s)}$
\EndProcedure
\end{algorithmic}
\end{algorithm}

For multiple sensors and targets, each target calibration $X^{(t)}$ must be checked individually.
Further, a sensor calibration $Y^{(s)}$ must only be adjusted if all targets detected by sensor $s$ are updated.
Due to the coupling of $X^{(t)}$ and $Y^{(s)}$ in the optimization, it is not possible that only a subset of all targets detected by a sensor is below the vehicle origin.

\section{Experiments}

In this section, our evaluation is described and the results are displayed and analyzed.
First, simulated data from~\cite{ali2019methods} is used in Section~\ref{sec:simulation} to compare our approach with existing HERW methods.
Afterward, experiments with real-world data of a robot and infrastructure cameras demonstrate the practicability of our proposed method in Section~\ref{sec:realWorldRobotics} and Section~\ref{sec:realWorldIntersection}.

If ground-truth calibrations are available, we use the error metric described in~\cite{horn2021online} defined by
\begin{subequations}
\label{eq:errorMetric}
\begin{alignat}{2}
    \varepsilon_r &= 2 \arccos(\vec{q}_{\varepsilon,1}) \pspace ,\\
    \varepsilon_t &= \norm{2 \, \quat{q}_{\varepsilon,d} \, \quat{q}^*_{\varepsilon,r}} \pspace,
\end{alignat}
\end{subequations}
with $\quat{q}_\varepsilon = \quat{q}_T^{-1} \, \sol{\quat{q}}$ and $\vec{q}_\varepsilon = \vectorize(\quat{q}_\varepsilon)$ since they describe physical entities.
All experiments were run on a computer with an ADM\,Ryzen\,7\,3700X CPU and 64GB of DDR4 RAM.
Whenever artificial noise is applied or execution times are determined, results are averaged over 100 runs.
In the following, our approach is compared to the methods of Dornaika et al.~\cite{dornaika1998simultaneous}, Shah et al.~\cite{shah2013solving}, Tabb et al.~\cite{tabb2017solving}, and Wang et al.~\cite{wang2022accurate}.
Furthermore, the HM and unit DQ approaches of Li et al.~\cite{li2010simultaneous} are evaluated.
They are referred to as \method{Dornaika}, \method{Li HM}, \method{Li DQ}, \method{Shah}, \method{Tabb}, and \method{Wang}, respectively.

\subsection{Simulation}
\label{sec:simulation}

The dataset \textsc{CS\_synthetic\_01} provided by the authors of~\cite{ali2019methods} contains 15 synthetic measurements of a robot arm and a checkerboard along with the ground-truth calibrations for both.
All measurement requirements mentioned in Section~\ref{sec:relWorkHandEye} are fulfilled.
Table~\ref{tab:ali} shows the HERW calibration results of different approaches.
For the lower half of Table~\ref{tab:ali}, data was augmented by applying zero-mean Gaussian noise with a standard derivation of \SI{1.0}{\centi\meter} and \SI{0.1}{\degree}, respectively.

\begin{table}[t]
    \centering
    \caption{%
        Results on CS\_synthetic\_01~\cite{ali2019methods}.
    }
    \label{tab:ali}
    \begin{threeparttable}
        \setlength{\tabcolsep}{0.5em}
        \begin{tabular}{clO{2.1}O{2.2}O{2.1}O{2.2}O{4.1}}

\toprule
& &
\multicolumn{2}{c}{$X$} &
\multicolumn{2}{c}{$Y$} &
\\

& \textbf{Method} &
$\varepsilon_t$ [$\si{\milli\metre}$] &
$\varepsilon_r$ [$\si{\degree}$] &
$\varepsilon_t$ [$\si{\milli\metre}$] &
$\varepsilon_r$ [$\si{\degree}$] &
{Time~[$\si{\milli\second}$]} \\

\midrule

\parbox[t]{1mm}{\multirow{6}{*}{\rotatebox[origin=c]{90}{original}}}
& \method{Dornaika}$^{*}$~\cite{dornaika1998simultaneous}  &  4.8  &  \B 0.01 &  8.3 & \B 0.02 &  32.4 \\
& \method{Li HM}~\cite{li2010simultaneous}                 &  \B 0.5  &  0.02 &  \B 2.4 & \B 0.02 & 0.4 \\
& \method{Li DQ}~\cite{li2010simultaneous}                 &  7.4  &  0.02 & 11.4 & \B 0.02 &  13.4 \\
& \method{Shah}~\cite{shah2013solving}                     &  7.5  &  0.04 &  8.9 & 0.05 & \B 0.3 \\
& \method{Tabb}$^{*}$~\cite{tabb2017solving}               &  3.1  &  0.02 &  3.9 & 0.03 & 195.3 \\
& \method{Wang}~\cite{wang2022accurate}                    &  7.5  &  0.04 &  8.9 & 0.05 & \B 0.3 \\
& \method{Ours}                                            &  8.8  &  0.02 & 13.3 & 0.03 &  16.6 \\

\midrule

\parbox[t]{1mm}{\multirow{6}{*}{\rotatebox[origin=c]{90}{noise}}}
& \method{Dornaika}$^{*}$~\cite{dornaika1998simultaneous}  & 35.2 & 0.11 &  35.9 & 0.11 &   37.6 \\
& \method{Li HM}~\cite{li2010simultaneous}                 & 37.3 & 0.12 &  38.1 & 0.12 &    0.4 \\
& \method{Li DQ}~\cite{li2010simultaneous}                 & 32.6 & 0.11 &  33.5 & 0.11 &   12.6 \\
& \method{Shah}~\cite{shah2013solving}                     & 36.8 & 0.15 &  37.2 & 0.15 & \B 0.3 \\
& \method{Tabb}$^{*}$~\cite{tabb2017solving}               & 37.5 & 0.11 &  37.0 & 0.11 &  268.9 \\
& \method{Wang}~\cite{wang2022accurate}                    & 36.8 & 0.15 &  37.2 & 0.15 & \B 0.3 \\
& \method{Ours}                                            & \B 31.2 & \B 0.10 &  \B 32.6 & \B 0.10 &   17.0 \\
 
\bottomrule
    
\end{tabular}

        \begin{tablenotes}\footnotesize
            \item $^{*}$ {Methods initialized with the results of \method{Li HM}~\cite{li2010simultaneous}}
        \end{tablenotes}
    \end{threeparttable}
\end{table}

In general, all approaches obtain solutions with rotation errors of similar magnitude on both original and noisy data.
Regarding the translation error on original data, \method{Li HM} is the most precise approach and achieves one of the lowest run times.
However, even with a relatively low amount of noise, \method{Li HM} degrades more than other methods.
As described in Section~\ref{sec:relatedWork}, it is known that linear methods like \method{Li HM} are less noise-robust than other methods.
All approaches find similar solutions on data with artificial noise, with slightly better results for our proposed method.
The error standard deviations for all methods are approximately \SI{15}{\milli\meter} and \SI{0.05}{\degree}.
Since our duality gap in this evaluation is always smaller than $10^{-8}$, we can certify the globality of our solution w.r.t. the cost function.

\subsection{Real-World Robot Data}
\label{sec:realWorldRobotics}

The dataset \textsc{kuka\_02} provided by the authors of~\cite{ali2019methods} contains 28 real-world measurements of a robot arm and a rigidly mounted camera with images of a checkerboard.
Since no ground-truth is available, the transformation cycle error $A_k \circ X \circ B_k\inv \circ Y\inv$ and the root-mean-square reprojection error $\varepsilon_\text{repr}$~\cite{ali2019methods} for all checkerboard detections are used for evaluation.

\begin{table}[t]
    \centering
    \caption{%
        Results on kuka\_02~\cite{ali2019methods}.
    }
    \label{tab:aliReal}
    \begin{threeparttable}
        \setlength{\tabcolsep}{0.5em}
        \begin{tabular}{lO{2.1}O{2.2}O{2.1}O{4.1}}

\toprule
&
\multicolumn{2}{c}{Cycle Error} &
\\

\textbf{Method} &
$\varepsilon_t$ [$\si{\milli\metre}$] &
$\varepsilon_r$ [$\si{\degree}$] &
$\varepsilon_\text{repr}$ [$\si{px}$] &
{Time~[$\si{\milli\second}$]} \\

\midrule

\method{Dornaika}$^{*}$~\cite{dornaika1998simultaneous}  &  \B 0.6  &  \B 0.04 &  1.04 & 58.3 \\
\method{Li HM}~\cite{li2010simultaneous}                 & 39.1  & \B 0.04 & 65.84 & 0.4 \\
\method{Li DQ}~\cite{li2010simultaneous}                 &  \B 0.6  & \B 0.04 & \B 0.98 & 23.4 \\
\method{Shah}~\cite{shah2013solving}                     &  \B 0.6  & \B 0.04 &  1.04 & 0.4 \\
\method{Tabb}$^{*}$~\cite{tabb2017solving}               &  0.7  &  0.05 &  1.56 & 34.4 \\
\method{Wang}~\cite{wang2022accurate}                    &  \B 0.6  & \B 0.04 &  1.04 & \B 0.3 \\
\method{Ours}                                            &  0.7  & \B 0.04 & \B 0.98 &  11.8 \\
 
\bottomrule
    
\end{tabular}

        \begin{tablenotes}\footnotesize
            \item $^{*}$ {Methods initialized with the results of \method{Li HM}~\cite{li2010simultaneous}}
        \end{tablenotes}
    \end{threeparttable}
\end{table}

The results on real-world data shown in Table~\ref{tab:aliReal} support the findings on simulated data in the previous section.
All methods except \method{Li HM} find similar solutions, while \method{Li HM} has by far the highest errors.
In addition, our approach has a verified globally optimal solution since the duality gap here is $10^{-11}$.
This shows that all approaches can calibrate the setup as long as the data fulfill all measurement requirements.
The following evaluation demonstrates the advantage of our method on planar-only data and with multiple targets and sensors.

\subsection{Real-World Intersection Data}
\label{sec:realWorldIntersection}

\begin{table*}[t]
    \caption{
        Results for the real-world intersection calibration.
    }
    \label{tab:realWorldInter}
    \begin{center}
    \begin{threeparttable}
        \def\arraystretch{1.0}
        \begin{tabular}{lcccc|O{3.1}O{1.2}|O{3.1}O{3.1}|O{3.1}O{3.1}|O{3.1}}

\toprule

& & & & &
\multicolumn{2}{c|}{\textbf{Checkerboard}} &
\multicolumn{2}{c|}{\textbf{Camera 1}} &
\multicolumn{2}{c|}{\textbf{Camera 2}} &
\\

\textbf{Method} & 
\multicolumn{2}{c}{\textbf{Targets}} &
\multicolumn{2}{c|}{\textbf{Sensors}} &
$\varepsilon_t$ [$\si{\centi\metre}$] &
$\varepsilon_r$ [$\si{\degree}$] &
$\varepsilon_\text{dist}$ [$\si{\%}$] &
$\varepsilon_\text{repr}$ [$\si{px}$] &
$\varepsilon_\text{dist}$ [$\si{\%}$] &
$\varepsilon_\text{repr}$ [$\si{px}$] &
{Time~[$\si{\milli\second}$]} \\

\midrule

\method{Li DQ}~\cite{li2010simultaneous}  &  CB  &       &  C1  &      &   804.0   &   0.47   &   106.9  &   14.5   &           &          &       18.1   \\
\method{Wang}~\cite{wang2022accurate}  &  CB  &       &  C1  &      &   220.9   &   2.17   &   31.1   &    14.7   &           &          &        0.3   \\
\method{Ours} (w/o $X$ norm)           &  CB  &       &  C1  &      &    35.1   &   \B 0.45   &    3.3   &    11.2   &           &          &       17.2   \\

\midrule

\method{Ours}                          &  CB  &       &  C1  &      &     2.5   &   0.49   &    \B 2.0   &   \B  5.0   &           &          &       22.8   \\
\method{Ours}                          &  CB  &       &      &  C2  &    11.9   &   0.49   &          &           &    1.6    &   22.0   &       22.9   \\
\method{Ours}                          &  CB  &       &  C1  &  C2  &     4.2   &   0.47   &    \B 2.0   &     5.1   &    \B 1.3    &    7.5   &       138.2   \\
\method{Ours}                          &  CB  &  AR   &  C1  &  C2  &     \B 2.2   &   0.83   &    2.4   &     7.8   &    1.7    &    \B 6.1   &      244.1   \\

\bottomrule

\end{tabular}

        \begin{tablenotes}
            \item The targets are the checkerboard (CB) and the front ArUco marker (AR).
            The sensors are both cameras C1 and C2.
        \end{tablenotes}
    \end{threeparttable}
    \end{center}
\end{table*}

We recorded real-world data using two infrastructure cameras with a resolution of $1920{\times}1200$\;\si{px}, located at the intersection described in~\cite{buchholz2022handling} with non-overlapping FOVs and a vehicle equipped with the targets described in Section~\ref{sec:calibTarget}.
Cropped images of both cameras are shown in Fig.~\ref{fig:vehicleImage}.
The target poses in camera coordinates were automatically estimated and the global vehicle poses were measured by an AMDA-G-Pro+ IMU with DGNSS.
In total, 134 checkerboard poses are available for the first camera and 33 for the second one.
For constraining the translation norm, the checkerboard distance to the vehicle origin was measured as~\SI{1.88}{\meter} using a folding rule.
The ground-truth checkerboard pose was measured with a laser rangefinder.
Since no ground-truth calibrations are available for the cameras, the root-mean-square reprojection error $\varepsilon_\text{repr}$~\cite{ali2019methods} of all checkerboard detections is used for evaluation.
Further, similar to~\cite{bartl2021automatic}, the distances between multiple ground points visible in each camera image were measured, as shown in Fig.~\ref{fig:vehicleImage}.
After calibration, the estimated camera poses and ground planes were used to project the respective image points onto the plane and estimate their distances.
The relative root-mean-square error between measured and estimated ground distances is denoted by~$\varepsilon_\text{dist}$.

Table~\ref{tab:realWorldInter} shows the results on infrastructure data.
The targets available for calibrating both cameras C1 and C2 are the checkerboard CB and the front ArUco marker AR.
Each row shows which targets and sensors were used for the respective calibration run.
Not all methods are shown since the behavior of all previously tested methods is similar when no translation norm is used.
As expected, a calibration without integrating a-priori knowledge of the translation norm is not feasible.
In contrast, integrating it into our approach leads to an accurate calibration for almost all tests.
Comparing the results for a separate calibration of C1 and C2 indicates that the amount of data recorded by C2 is insufficient w.r.t. the amount of noise within the data.

We have also used the ground-truth calibration of the checkerboard for comparison with PnP on camera 1, yielding $\varepsilon_\text{dist}$ of \SI{2.0}{\%} and $\varepsilon_\text{repr}$ of \SI{3.9}{px}.
Compared to our results of \SI{2.0}{\%} and \SI{5.0}{px}, the effort of measuring the exact checkerboard pose for enabling PnP does not bring a significant benefit, especially in the ground distance metric.

For the joint calibration of C1 and C2, our multi-sensor approach compensates for the previously mentioned lack of data for C2, which helps to improve the result for C2 without compromising the calibration of C1.
Since our method is designed for geo-referenced calibration only, it is impossible to compare our results with other approaches for local-referenced infrastructure calibration~\cite{sochor2017traffic, bhardwaj2018autocalib, bartl2021automatic}.
However, with $\varepsilon_\text{dist}$ of \SI{2.0}{\%} and \SI{1.3}{\%}, we achieve promising results regarding other state-of-the-art methods with values from \SI{3}{\%} to \SI{9}{\%}.

Although the ArUco marker pose estimation is less accurate than the checkerboard pose estimation, using AR partially improves the results for C2 compared to the single-target approach.
However, the results for C1 slightly degrade in this case.
Nevertheless, using multiple checkerboard targets instead of ArUco markers would presumably further improve the calibration results.

\section{Conclusion}

We have described a QCQP approach for solving the HERW problem certifiably globally optimal, supporting multiple sensors and targets.
The evaluation shows that our approach can obtain state-of-the-art results while maintaining sub-second execution times, even for multiple sensors and targets. 
Furthermore, our approach can incorporate a-priori knowledge, i.e., the translation norm of a target calibration.
This enables calibration using planar-only pose data provided by infrastructure sensors, even though the measurement requirements for a unique solution are not fulfilled in this case.
Of all compared approaches, our method is the only one providing accurate results on planar-only pose data.

We have shown the infrastructure calibration only for cameras, using a checkerboard as calibration target.
However, this concept can also be used for other sensor types, e.g., with calibration targets like \cite{guindel2017automatic} for lidars.

\bibliographystyle{IEEEtran}
\bibliography{IEEEabrv,mybibfile}

\begin{thebibliography}{10}
\providecommand{\url}[1]{#1}
\csname url@rmstyle\endcsname
\providecommand{\newblock}{\relax}
\providecommand{\bibinfo}[2]{#2}
\providecommand\BIBentrySTDinterwordspacing{\spaceskip=0pt\relax}
\providecommand\BIBentryALTinterwordstretchfactor{4}
\providecommand\BIBentryALTinterwordspacing{\spaceskip=\fontdimen2\font plus
\BIBentryALTinterwordstretchfactor\fontdimen3\font minus
  \fontdimen4\font\relax}
\providecommand\BIBforeignlanguage[2]{{%
\expandafter\ifx\csname l@#1\endcsname\relax
\typeout{** WARNING: IEEEtran.bst: No hyphenation pattern has been}%
\typeout{** loaded for the language `#1'. Using the pattern for}%
\typeout{** the default language instead.}%
\else
\language=\csname l@#1\endcsname
\fi
#2}}

\bibitem{buchholz2022handling}
M.~Buchholz, J.~Müller, M.~Herrmann, J.~Strohbeck, B.~Völz, M.~Maier,
  J.~Paczia, O.~Stein, H.~Rehborn, and R.-W. Henn, ``Handling occlusions in
  automated driving using a multiaccess edge computing server-based environment
  model from infrastructure sensors,'' \emph{IEEE Intelligent Transportation
  Systems Magazine}, vol.~14, no.~3, pp. 106--120, 2022.

\bibitem{datondji2016survey}
S.~R.~E. Datondji, Y.~Dupuis, P.~Subirats, and P.~Vasseur, ``A survey of
  vision-based traffic monitoring of road intersections,'' \emph{IEEE
  Transactions on Intelligent Transportation Systems}, vol.~17, no.~10, pp.
  2681--2698, 2016.

\bibitem{marchand2016pose}
E.~Marchand, H.~Uchiyama, and F.~Spindler, ``{Pose Estimation for Augmented
  Reality: A Hands-On Survey},'' \emph{{IEEE Transactions on Visualization and
  Computer Graphics}}, vol.~22, no.~12, pp. 2633 -- 2651, 2016.

\bibitem{tsai1988real}
R.~Y. Tsai and R.~K. Lenz, ``Real time versatile robotics hand/eye calibration
  using 3d machine vision,'' in \emph{IEEE International Conference on Robotics
  and Automation (ICRA)}, 1988, pp. 554--561.

\bibitem{zhuang1994simultaneous}
H.~Zhuang, Z.~Roth, and R.~Sudhakar, ``Simultaneous robot/world and tool/flange
  calibration by solving homogeneous transformation equations of the form
  ax=yb,'' \emph{IEEE Transactions on Robotics and Automation}, vol.~10, no.~4,
  pp. 549--554, 1994.

\bibitem{tabb2017solving}
A.~Tabb and K.~M. Ahmad~Yousef, ``Solving the robot-world hand-eye(s)
  calibration problem with iterative methods,'' \emph{Machine Vision and
  Applications}, vol.~28, no.~5, pp. 569--590, 2017.

\bibitem{heller2014handeye}
J.~Heller, D.~Henrion, and T.~Pajdla, ``Hand-eye and robot-world calibration by
  global polynomial optimization,'' in \emph{IEEE International Conference on
  Robotics and Automation (ICRA)}, 9 2014, pp. 3157--3164.

\bibitem{horn2021online}
M.~Horn, T.~Wodtko, M.~Buchholz, and K.~Dietmayer, ``Online extrinsic
  calibration based on per-sensor ego-motion using dual quaternions,''
  \emph{IEEE Robotics and Automation Letters}, vol.~6, no.~2, pp. 982--989,
  2021.

\bibitem{li2018eye}
Z.~Li and V.~Willert, ``Eye-to-eye calibration for cameras with disjoint fields
  of view,'' in \emph{IEEE International Conference on Intelligent
  Transportation Systems (ITSC)}, 2018, pp. 2631--2638.

\bibitem{ali2019methods}
I.~Ali, O.~Suominen, A.~Gotchev, and E.~R. Morales, ``Methods for simultaneous
  robot-world-hand--eye calibration: A comparative study,'' \emph{Sensors},
  vol.~19, no.~12, 2019.

\bibitem{dornaika1998simultaneous}
F.~Dornaika and R.~Horaud, ``Simultaneous robot-world and hand-eye
  calibration,'' \emph{IEEE Transactions on Robotics and Automation}, vol.~14,
  pp. 617--622, 1998.

\bibitem{shah2013solving}
M.~Shah, ``Solving the robot-world/hand-eye calibration problem using the
  kronecker product,'' \emph{Journal of Mechanisms and Robotics}, vol.~5,
  no.~3, 2013.

\bibitem{wang2022accurate}
Y.~Wang, W.~Jiang, K.~Huang, S.~Schwertfeger, and L.~Kneip, ``Accurate
  calibration of multi-perspective cameras from a generalization of the
  hand-eye constraint,'' in \emph{IEEE International Conference on Robotics and
  Automation (ICRA)}, 5 2022, pp. 1244--1250.

\bibitem{li2010simultaneous}
A.~Li, L.~Wang, and D.~Wu, ``Simultaneous robot-world and hand-eye calibration
  using dual-quaternions and kronecker product,'' \emph{International Journal
  of Physical Sciences}, vol.~5, no.~10, pp. 1530--1536, 2010.

\bibitem{li2018simultaneous}
W.~Li, M.~Dong, N.~Lu, X.~Lou, and P.~Sun, ``Simultaneous robot--world and
  hand--eye calibration without a calibration object,'' \emph{Sensors},
  vol.~18, no.~11, 2018.

\bibitem{zhao2019simultaneous}
Z.~Zhao, ``Simultaneous robot-world and hand-eye calibration by the alternative
  linear programming,'' \emph{Pattern Recognition Letters}, vol. 127, pp.
  174--180, 2019.

\bibitem{kanhere2010taxonomy}
N.~K. Kanhere and S.~T. Birchfield, ``A taxonomy and analysis of camera
  calibration methods for traffic monitoring applications,'' \emph{IEEE
  Transactions on Intelligent Transportation Systems}, vol.~11, no.~2, pp.
  441--452, 2010.

\bibitem{fung2003camera}
G.~S.~K. Fung, N.~H.~C. Yung, and G.~K. Pang, ``Camera calibration from road
  lane markings,'' \emph{Optical Engineering}, vol.~42, pp. 2967--2977, 2003.

\bibitem{dubska2014automatic}
M.~Dubsk{\'a}, J.~Sochor, and A.~Herout, ``Automatic camera calibration for
  traffic understanding,'' in \emph{British Machine Vision Conference}, 2014.

\bibitem{dubska2015fully}
M.~Dubsk{\'a}, A.~Herout, R.~Jur{\'a}nek, and J.~Sochor, ``Fully automatic
  roadside camera calibration for traffic surveillance,'' \emph{IEEE
  Transactions on Intelligent Transportation Systems}, vol.~16, no.~3, pp.
  1162--1171, 2015.

\bibitem{sochor2017traffic}
J.~Sochor, R.~Juránek, and A.~Herout, ``Traffic surveillance camera
  calibration by 3d model bounding box alignment for accurate vehicle speed
  measurement,'' \emph{Computer Vision and Image Understanding}, vol. 161, pp.
  87--98, 2017.

\bibitem{bhardwaj2018autocalib}
R.~Bhardwaj, G.~K. Tummala, G.~Ramalingam, R.~Ramjee, and P.~Sinha,
  ``Autocalib: Automatic traffic camera calibration at scale,'' \emph{ACM
  Transactions on Sensor Networks}, vol.~14, 11 2018.

\bibitem{bartl2021automatic}
V.~Bartl, J.~{\v{S}}pa{\v{n}}hel, P.~Dobe{\v{s}}, R.~Jur{\'a}nek, and
  A.~Herout, ``Automatic camera calibration by landmarks on rigid objects,''
  \emph{Machine Vision and Applications}, vol.~32, no.~1, pp. 1--13, 2021.

\bibitem{mueller2019laci}
J.~Müller, M.~Herrmann, J.~Strohbeck, V.~Belagiannis, and M.~Buchholz, ``Laci:
  Low-effort automatic calibration of infrastructure sensors,'' in \emph{IEEE
  Intelligent Transportation Systems Conference (ITSC)}, 2019, pp. 3928--3933.

\bibitem{tsaregorodtsev2022extrinsic}
A.~Tsaregorodtsev, J.~Müller, J.~Strohbeck, M.~Herrmann, M.~Buchholz, and
  V.~Belagiannis, ``Extrinsic camera calibration with semantic segmentation,''
  in \emph{International Conference on Intelligent Transportation Systems
  (ITSC)}, 2022, pp. 3781--3787.

\bibitem{mccarthy1990introduction}
J.~M. McCarthy, \emph{Introduction to Theoretical Kinematics}.\hskip 1em plus
  0.5em minus 0.4em\relax MIT Press, 1990.

\bibitem{wodtko2021globally}
T.~Wodtko, M.~Horn, M.~Buchholz, and K.~Dietmayer, ``Globally optimal
  multi-scale monocular hand-eye calibration using dual quaternions,'' in
  \emph{International Conference on 3D Vision (3DV)}, 2021, pp. 249--257.

\bibitem{boyd2004convex}
S.~P. Boyd and L.~Vandenberghe, \emph{Convex optimization}.\hskip 1em plus
  0.5em minus 0.4em\relax Cambridge University Press, 2004.

\bibitem{daniilidis1999hand}
K.~Daniilidis, ``Hand-eye calibration using dual quaternions,'' \emph{The
  International Journal of Robotics Research}, vol.~18, no.~3, pp. 286--298,
  1999.

\bibitem{fischler1981random}
M.~A. Fischler and R.~C. Bolles, ``Random sample consensus: A paradigm for
  model fitting with applications to image analysis and automated
  cartography,'' \emph{Commun. ACM}, vol.~24, pp. 381--395, 6 1981.

\bibitem{garrido2014automatic}
S.~Garrido-Jurado, R.~Mu{\~n}oz-Salinas, F.~J. Madrid-Cuevas, and M.~J.
  Mar{\'\i}n-Jim{\'e}nez, ``Automatic generation and detection of highly
  reliable fiducial markers under occlusion,'' \emph{Pattern Recognition},
  vol.~47, no.~6, pp. 2280--2292, 2014.

\bibitem{furgale2013unified}
P.~Furgale, J.~Rehder, and R.~Siegwart, ``Unified temporal and spatial
  calibration for multi-sensor systems,'' in \emph{IEEE/RSJ International
  Conference on Intelligent Robots and Systems (IROS)}, 2013, pp. 1280--1286.

\bibitem{guindel2017automatic}
C.~Guindel, J.~Beltrán, D.~Martín, and F.~García, ``Automatic extrinsic
  calibration for lidar-stereo vehicle sensor setups,'' in \emph{IEEE
  International Conference on Intelligent Transportation Systems (ITSC)}, 2017,
  pp. 1--6.

\end{thebibliography}




\end{document}